# Artificial Intelligence for Green Hydrogen Yield Prediction and Site Suitability using SHAP-Based Composite Index: Focus on Oman

Obumneme Zimuzor Nwafor[1] and Mohammed Abdul Majeed Al Hooti[2]

[1]School of Computing, Engineering and Built Environment,
Glasgow Caledonian University, Scotland
ORCID: 0000-0002-0993-1659.

[2] Department of Business Management, University of Lancashire United Kingdom
 ORCID: 0009-0005-5793-7122

## Abstract
As nations seek sustainable alternatives to fossil fuels, green hydrogen has emerged as a promising strategic pathway toward decarbonisation, particularly in solar-rich arid regions. However, identifying optimal locations for hydrogen production requires the integration of complex environmental, atmospheric, and infrastructural factors, often compounded by limited availability of direct hydrogen yield data. This study presents a novel Artificial Intelligence (AI) framework for computing green hydrogen yield and site suitability index using mean absolute SHAP (SHapley Additive exPlanations) values. This framework consists of a multi-stage pipeline of unsupervised multi-variable clustering, supervised machine learning classifier and SHAP algorithm.  The pipeline trains on an integrated meteorological, topographic and temporal dataset and the results revealed distinct spatial patterns of suitability and relative influence of the variables. With model predictive accuracy of 98%, the result also showed that water proximity, elevation and seasonal variation are the most influential factors determining green hydrogen site suitability in Oman with mean absolute shap values of 2.470891, 2.376296 and 1.273216 respectively. Given limited or absence of ground-truth yield data in many countries that have green hydrogen prospects and ambitions, this study offers an objective and reproducible alternative to subjective expert weightings, thus allowing the data to speak for itself and potentially discover novel latent groupings without pre-imposed assumptions. This study offers industry stakeholders and policymakers a replicable and scalable tool for green hydrogen infrastructure planning and other decision making in data-scarce regions.

# 1. Introduction
Green hydrogen has emerged as an important enabler in the global transition to low-carbon economies because it offers a clean energy alternative for power generation, industrial processes, and transportation. Produced through renewable-powered electrolysis using solar and wind energy, it serves as a zero-emission alternative to conventional fossil-based fuels. However, identifying optimal sites for green hydrogen production presents both technical and policy challenges due to the multi-dimensionality and uncertainty of interconnected factors. As

---

**Corresponding author**: Obumneme Nwafor obumnemenwafor@gmail.com

countries pursue hydrogen strategies to meet their net-zero targets, arid and semi-arid regions have attracted increasing attention because of their unique potentials in meteorological and topographical factors, in addition to strategic proximity to international energy markets. With one of the world's highest solar energy potentials, emerging desalination infrastructure, and a strong policy commitment to renewable energy, Oman is being considered as a future hub for green hydrogen production and export [1]. With exceptional renewable energy resources including solar irradiation of up to 2500 kWh/m² and wind speeds up to 8.3 m/s, combined with proactive government initiatives for the Hydrogen economy, the country is an ideal case study for green hydrogen development [2]. Furthermore, it is strategically located with over 10% of all global trades passing through the Strait of Hormuz and Bab El-Mandeb, and over 40% of global container capacity passing though the Suez Canal [3]. However, transitioning from national ambition to practical implementation requires a robust framework for site selection, yield forecasting, and investment prioritization, particularly in the absence of granular hydrogen production data. Identifying optimal sites for green hydrogen production is inherently a multi-criteria spatial decision-making problem. Earlier studies in this domain have relied heavily on expert-driven Multi-Criteria Decision Analysis (MCDA) methods, such as the Analytic Hierarchy Process (AHP) and Fuzzy Logic, to evaluate site suitability for renewable energy projects such as green hydrogen [4] [5]. While these approaches are useful, they are often vulnerable to subjectivity due to static weight assignments and may fail to capture complex feature interactions or nonlinear effects particularly in new data. Moreover, in many emerging hydrogen economies, direct measurements of hydrogen yield or techno-economic performance are limited, hence the need for proxy-driven models [6].

## 2. Related Literature

Many Gulf countries including Oman are rapidly positioning themselves as hubs for green hydrogen production due to abundant solar and other meteorological resources in the region [7]. Multiple studies have highlighted Oman's favourable conditions for green hydrogen production, including vast desert landscapes, low population density, and proximity to maritime trade routes [8] [9]. [10] further reveals that 99.57% of planned capacity additions in Oman are expected to be based on solar and wind, with over 80% directly linked to green hydrogen production. Also, [11] highlights the economic viability of green hydrogen in the region and [12] evaluated hybrid PV/wind systems in Dhofar, finding an average cost of hydrogen (LCOH) of $4.65/kg with over 85% renewable energy fraction. [10] also estimates the LCOH powered by solar PV at $5.63/kg, which is very competitive for global export markets. Some studies such as [14] and [15] highlight methods of site selection using AHP in which experts assign weights to various factors and aggregate them into a composite production yield index. While MCDA approaches are relatively easy to implement, they have been criticized for subjectivity, lack of adaptability, and inability to model nonlinear relationships [16]. In data-scarce regions where expert judgments dominate, this can lead to inconsistencies and reproducibility issues [17]. Recent studies have also applied various ML models to predict hydrogen production potential based on environmental and operational data [18][19]. For instance, [20] employed geological modelling with open-source data for natural hydrogen mapping, and [18] introduced a fuzzy-decision model for hydrogen refuelling stations [21] simulated a 7 GW solar-powered green hydrogen supply chain for residential cooking, and [22] conducted a location-based economic comparison across 15 sites in Oman.

## 2.1 Research Gap and Novel Contribution of the Current Study

Despite increasing momentum and research in green hydrogen initiatives, a significant gap persists across both emerging and developed economies. In the case of Oman's growing ambition to become a global hydrogen hub, there is limited availability of ground-truth data, which hinders accurate assessment of site-specific production potential [23]. A similar Gulf-wide analyses reveal informational deficiencies due to absence of data transparency hinder effective planning for Green Hydrogen [24]. Also, [25] emphasises this concern but noting that Oman's aspiration in hydrogen needs improved data infrastructures to support large-scale planning and investment. These gaps manifest as a lack of standardized yield models, missing environmental performance data, and unclear metrics for site prioritization. This challenge is not unique to the Gulf region; even in developed contexts, a 2023 UK government report identifies poor data availability, particularly yield potential as a major bottleneck for local hydrogen planning [26].

In the light of limited hydrogen yield data, this study adopts an unsupervised classification approach using meteorological, topographic and temporal features to group locations by their latent suitability characteristics. This method provides a reproducible and objective alternative to expert-based scoring in the absence of direct yield observations. This clustering technique is combined with a SHAP-guided AI framework to compute green hydrogen yield index and site suitability. SHAP is grounded in cooperative game theory and provides consistent local and global explanations for model predictions [27]. It has been successfully applied in healthcare [28], climate modelling [29], and energy forecasting [30] as a transparent mechanism for understanding complex models. Few studies have used SHAP to dynamically inform factor weightings in decision-support systems, and almost none have extended this approach to green hydrogen yield prediction in arid and data-scarce contexts. To our knowledge, this is the first AI-based green hydrogen yield and location suitability study that uses explainable AI to determine weights for composite index construction.

## 2.2 Justification for Using Unsupervised Proxy Classification

Using unsupervised learning to derive proxy suitability classes is often a viable approach when there is limited or no ground truth data [31]. In many geospatial planning problems, there's no "true" label of site suitability unless physical field surveys or economic feasibility studies were done. In such instance, clustering techniques can group similar data points into meaningful classes based on patterns in the data itself. This is especially useful when target variable is a latent construct influenced by multiple interacting factors [33]. This allows the data to speak instead of pre-imposing assumptions (e.g., solar = 40%, water = 30%). This can reveal novel groupings not obvious from literature-based weighting schemes. Furthermore, this approach is objective and repeatable because unlike expert-based weights (which can be subjective and differ by source), clustering is fully reproducible and based on a mathematical formulation [34].

# 3. Methodology

## 3.1 The Dataset

The dataset for this study was curated by integrating multi-source satellite and meteorological data using a customized pipeline developed in Python and Google Earth Engine (GEE). The region of interest (ROI) consists of ten cities across the Sultanate of Oman namely Muscat, Salalah, Duqm, Sohar, Sur, Nizwa, Ibri, Ibra, Khasab, and Al Jazer selected for their strategic relevance to renewable energy development and geographic diversity. For each location, daily meteorological data such as including solar irradiance, temperature, and wind speed was obtained from NASA's Prediction of Worldwide Energy Resources (POWER) project. Also, atmospheric aerosol conditions were captured using satellite-derived Aerosol Optical Depth (AOD) measurements from the Moderate Resolution Imaging Spectroradiometer (MODIS) and European Space Agency's Sentinel-5P Tropospheric Monitoring Instrument (TROPOMI) collections, filtered to prioritize the 0.47 μm spectral band most sensitive to fine-mode particulate matter. AOD values were spatially averaged over a 5 km buffer radius centred on the coordinates of each city and temporally aligned with the meteorological records from NASA POWER. Additional geospatial features such as land cover classification and proximity to surface water were extracted using the ESA WorldCover and Global Surface Water Explorer datasets, respectively. Temporal interpolation techniques were applied to fill missing AOD records, and all features were normalized to enable cross-location comparison. The resulting dataset captures a high-resolution, multi-feature representation of Oman's environmental and atmospheric conditions relevant to green hydrogen production, covering the period from January 2020 to December 2024. Table 1 below shows a list and description of the variables in the dataset.

**Table 1.** Dataset Feature Names, Units and Description

| Feature Name | Unit / Data Type | Description | Relevance(why it matters) |
|---|---|---|---|
| Solar Irradiance | W/m² (Watts per square meter) | Measures the power of solar radiation received per unit area. | Critical for solar-powered energy in green hydrogen systems; high irradiance improves solar PV/electrolysis efficiency. |
| Temperature | °C (Degrees Celsius) | Daily average or maximum air temperature at 2 meters above ground. | Influences electrolysis efficiency, water evaporation rates, and cooling requirements for equipment |
| Wind Speed | m/s (Meters per second) | Near-surface wind speed, typically at 10 meters elevation. | Enables integration of wind energy into hybrid renewable hydrogen systems; affects turbine output and feasibility of off-grid operations. |
| AOD (Aerosol Optical Depth) | Unitless (dimensionless) | Represents the degree to which aerosols prevent light transmission; values typically range from 0 to >3. | High AOD reduces solar irradiance, affecting hydrogen production yield and system efficiency; also indicates potential maintenance due to dust accumulation. |
| Land Cover Class | Integer (categorical code) | Coded land classification by ESA WorldCover (e.g 10 = cropland, 20 = forest, etc.). | Determines land availability, suitability, and regulatory constraints for hydrogen infrastructure deployment (e.g., avoid protected forests or farmlands). |
| Water Proximity | Kilometers (km) | Euclidean distance from each site to nearest surface water body. | Important for electrolysis-based hydrogen production, Operational cost and infrastructure cooling |
| Elevation | Meters (m) above sea level | Ground elevation at the location based on digital elevation model (DEM). | Affects temperature, air density, and water transport energy costs |
| Month | Unitless (dimensionless) | Month of the year | Captures seasonal variability in irradiance, temperature, and AOD; enables modelling of temporal effects on hydrogen yield and resource planning. |

### 3.2 Machine Learning Algorithm Pipeline

This study adopts a four-stage methodology comprising the following:

i. Unsupervised multi-variable clustering algorithm to generate a proxy target yield class.
ii. Supervised classification models to learn patterns between input features and the proxy classes;
iii. SHAP explainability to determine the global variable importance ranking;

iv. And a composite site suitability index (SCI), computed using mean absolute SHAP values as variable weights.

Figure 1 below shows a schematic diagram of the experiment methodology for study.

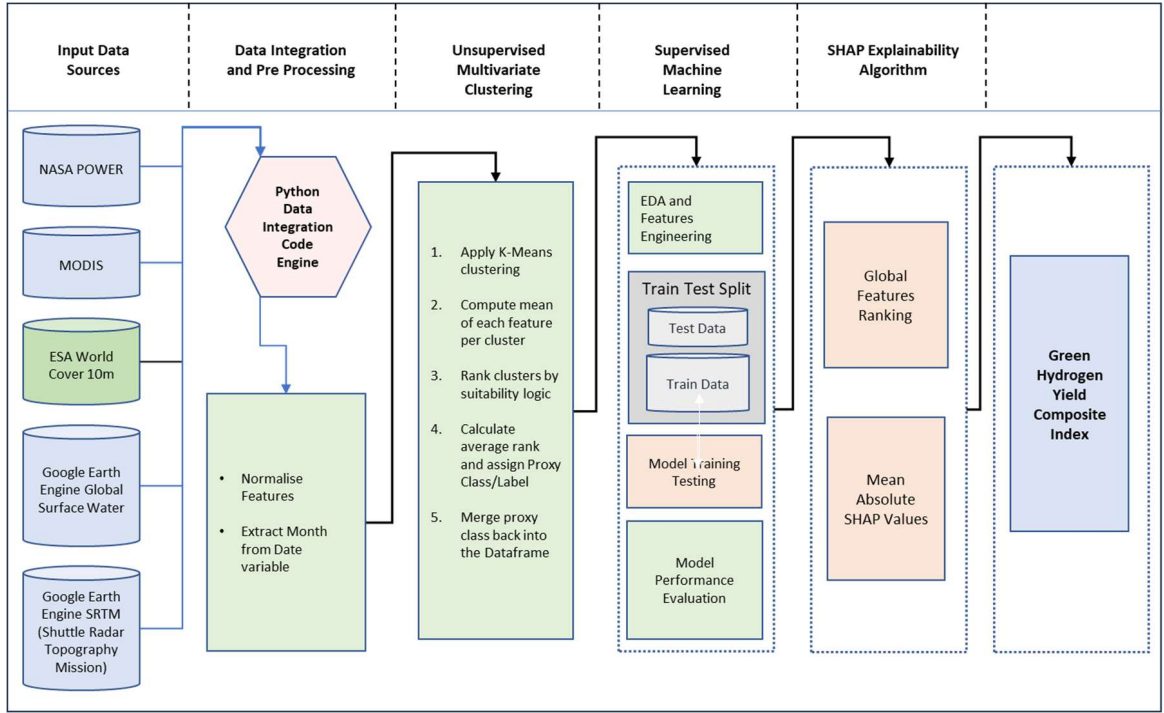

**Figure 1** Schematic diagram of the experiment methodology for using unsupervised, supervised and explainable AI pipeline for composite index yield prediction and location suitability for green hydrogen production.

**Stage 1: Unsupervised Multi-Variable Cluster Ranking for Proxy Target Generation**

The first step of this pipeline involves the use of an unsupervised clustering approach to derive proxy target classes for production yield and site suitability. A multivariate dataset was normalised using the MinMax Scaler represented by Equation (1) below:

$$\hat{X}_{ij} = \frac{X_{ij} - \min(X_{.j})}{\max(X_{.j}) - \min(X_{.j})} \quad (1)$$

Where $X_{ij}$ is the original value of feature j for day I, $\min(X_{.j})$ is the minimum value of feature j across the entire training set, $\max(X_{.j})$ represents the maximum value of feature j across the entire training set and $\hat{X}_{ij}$ is the scaled value constrained to the interval [0,1].

The normalised features serve as input for unsupervised clustering using the K-Means algorithm, where the optimal number of clusters, K, was determined using both the Elbow Method and Silhouette Score [35][36]. Each data point $x_i \in R^n$ is assigned to a cluster $C_k$ such that the within-cluster sum of squares is minimized:

$$\arg\min \sum_{K=1}^{K} \sum_{x_i \in C_k} \|x_i - \mu_k\|^2 \quad (2)$$

Where $\mu_k$ is the centroid of cluster $C_k$. After clustering, the resulting clusters are ranked in turn into proxy classes {0: "Very Low", 1: "Low", 2: "Moderate", 3: "High", 4: "Very High"} based on aggregate feature statistics of each of the variables. This approach aligns with recent studies employing clustering to compensate for data-scarce regions in spatial energy planning [37] [38].

**Stage 2: Supervised Classification of Proxy Classes**

Following clustering, a supervised machine learning classifier is trained on the original input variables, using the derived proxy classes as the target variables. The study utilizes the Extreme Gradient Boosting (XGBoost) classifier, known for its robustness to multi-collinearity, handling of missing data, and high predictive performance [19]. The classifier learns a non-linear mapping:

$$\hat{y}_i = f(x_i) = \sum_{m=1}^{M} f_m(x_i), \ f_m \in \varphi \qquad (3)$$

Where $\varphi$ denotes the space of decision trees, M is the number of boosted trees and the target variable $\hat{y}_i$ represents the predicted proxy class. The model is trained using gradient descent to minimize a regularized objective function that balances classification loss and model complexity. Cross-validation and confusion matrix analysis are used to assess classification accuracy. This stage ensures that the learned suitability signal is generalizable across spatial locations, enabling data-driven predictions in regions with similar conditions.

**Stage 3: SHAP-Based Explainability and Feature Importance Ranking**

To enhance transparency and interpretability, SHAP values are computed for the trained classifier. SHAP assigns each feature an importance value representing its average contribution to the prediction across all samples, based on Shapley values from cooperative game theory [39]. The SHAP value $\emptyset_j$ for feature j is defined as:

$$\emptyset_j = \sum_{S \subseteq F\{j\}} \frac{|S|!(|F|-|S|-1)!}{|F|!} [f_{S \cup \{j\}}(x) - f_S(x)] \qquad (4)$$

Where F is the set of all features and S represents a subset of features excluding j. The mean absolute SHAP value for each feature is calculated to determine its global influence on model predictions. This data-driven feature importance ranking replaces subjective expert weighting traditionally used in Multi-Criteria Decision Analysis (MCDA), offering a principled and reproducible approach to index construction [40].

**Stage 4: Composite Index Construction**

The final stage constructs a Composite Yield and Site Suitability Index using the SHAP-derived feature importance as variable weights. For each site i, the index is calculated as a weighted sum of normalized feature values:

$$SCI_i = \sum_{j=1}^{n} w_j \cdot \hat{X}_{ij} \qquad (5)$$

Where

$$w_j = \frac{|\emptyset_j|}{\sum_{k=1}^{n} |\emptyset_k|} \qquad (6)$$

Here, $\hat{X}_{ij}$ is the normalized (and directionally adjusted, if applicable) value of feature j for site iii, and $w_j$ is the normalized mean absolute SHAP value for that feature. This formulation ensures that features with higher explanatory power in the model receive greater influence in the composite index, thereby aligning the index with actual data-driven impacts rather than static or heuristic weights. This novel use of SHAP-guided weighting in site suitability indexing enhances transparency, scalability, and interpretability—making it especially useful for national energy planning in data-scarce or rapidly evolving contexts [41].

## 4. Results and Discussion

### 4.1 Exploratory Data Analysis

Extensive exploratory data analysis (EDA) was conducted to provide insights into the distributions, relationships, and temporal patterns among the variables. First, histograms with kernel density estimates (KDE) were generated for solar irradiance, temperature, and wind speed as shown in Figure 2.

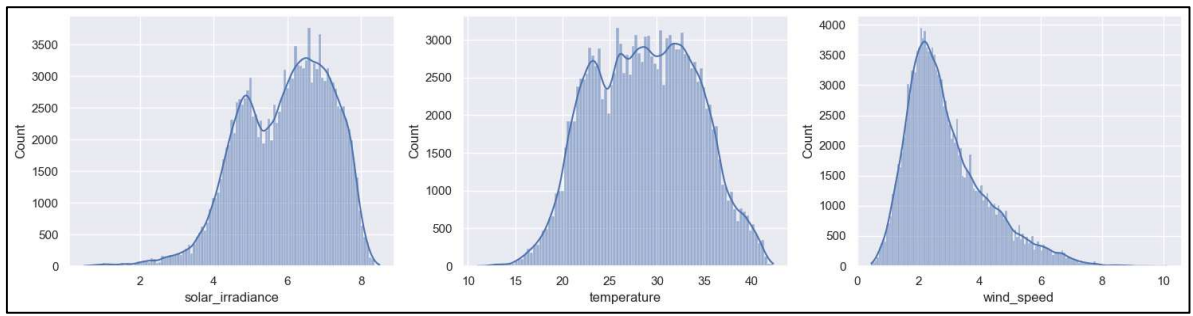

**Figure 2** Distribution plots of Solar Irradiance, Temperature and Wind Speed showing kernel density estimation (KDE) curves

Solar irradiance demonstrates a bimodal distribution with peaks around 5.0 and 6.5 kWh/m²/day, a mean of 5.71 kWh/m²/day, and a moderate right skewness of 0.42, indicative of seasonal variation influenced by cloud cover and dust load [42] [43]. The temperature distribution is approximately uniform, ranging from 15°C to 42°C, with a mean of 27.4°C and low skewness (−0.11), consistent with the hot desert climate (BWh) classification of Oman under the Köppen–Geiger system [44][45]. This thermal stability is advantageous for photovoltaic (PV) efficiency, which tends to decline under extreme heat stress [46]. Wind speed exhibits a positively skewed distribution (skewness = 1.23), with most values clustered between 1.5 and 4.0 m/s, and a mean of 3.18 m/s, highlighting the intermittent nature of wind resources in the region. While solar potential is strong and consistent, wind resources appear modest and location-specific. This shows that there is a need for hybrid system optimisation and geospatial prioritisation in green hydrogen infrastructure planning [47] [48]. These findings show the importance of data-driven site assessment that accounts for intra-annual variation and resource co-occurrence.

The monthly AOD trends across the nine cities in Oman used in this study is plotted as show in Figure 3 below. The plot shows that AOD levels exhibit a clear seasonal peak during the summer months particularly in June and July which coincides with the regional dust season driven by intensified north-westerly winds, convective activity, and subsidence over the Arabian Peninsula [49] [50]. Duqm recorded the highest peak AOD in July (over 0.61), followed by Sur and Muscat, indicating significant dust loading in coastal and central-eastern zones. In contrast, inland cities like Ibri and Nizwa consistently reported lower AOD levels, ranging between 0.15 and 0.35 across most months, suggesting relatively lower dust exposure. The intra-annual variation reveals both synoptic-scale atmospheric circulation and local topographic shielding [51]. Mean annual AOD values across cities ranged from 0.22 (Salalah) to 0.36 (Duqm), aligning with prior satellite-based observations over Oman [52] [53].

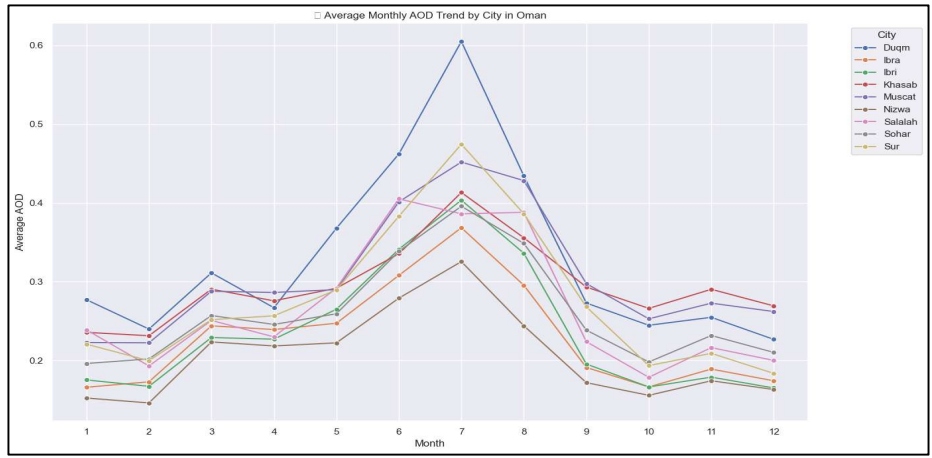

**Figure 3** Monthly AOD trends across the nine cities in region of interest (Oman) used

These findings highlight the necessity of integrating city-specific aerosol profiles into the green hydrogen yield assessments given that varying levels of AOD affect the photovoltaic efficiency of solar-powered green hydrogen plants [54] [55]. Consequently, strategic site selection for green hydrogen infrastructure must consider aerosol seasonality and dust mitigation strategies to optimise system yield. The correlation matrix presented in Figure 4 provides a comprehensive overview of the linear relationships among variables in the dataset.

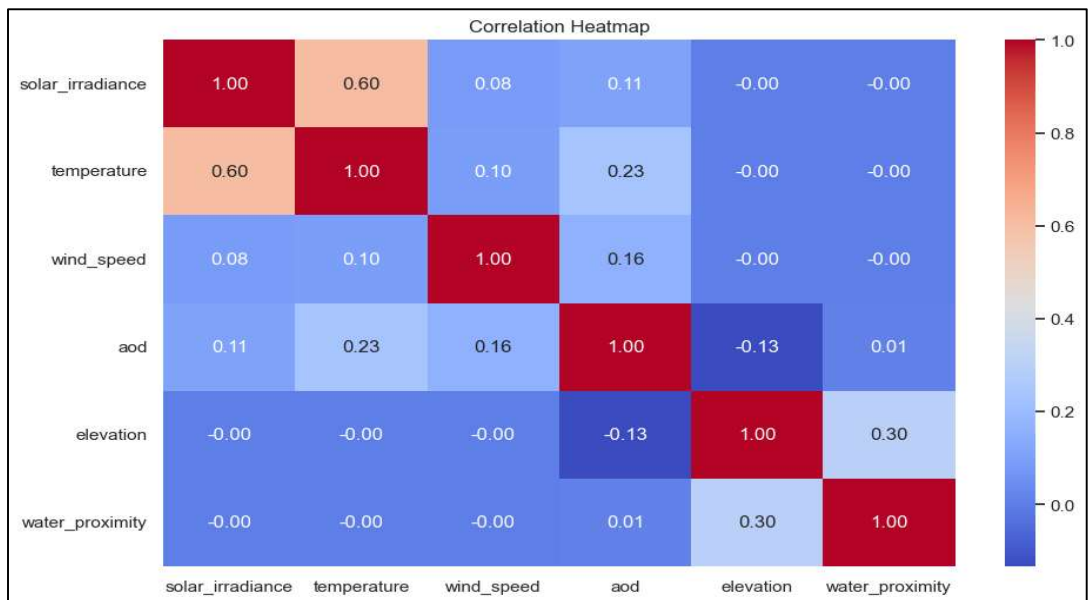

**Figure 4** Correlation Matrix of the Input Features for a combined dataset of customer consumption and Staff operational activities. The correlation coefficient has a bound of {1,-1} with Pearson Coefficient r scale zones defined as {High: $0.7 < |r| \leq 1.0$, Moderate:

As expected, strong positive correlation (r = 0.60) is observed between solar irradiance and temperature, shows the relationship between solar resource availability and thermal potential, both of which are beneficial for photovoltaic and thermochemical hydrogen systems. In contrast, wind speed shows only weak positive correlations with solar irradiance (r = 0.08) and temperature (r = 0.10), suggesting temporal or locational decoupling between the variables. This should be taken into account in hybrid system design to ensure complementary resource coverage. AOD exhibits a modest positive correlation with temperature (r = 0.23) and wind speed (r = 0.16), aligning with literature that attributes aerosol mobilization to increased convective uplift and surface winds during warmer periods [56]. However, AOD's correlation with solar irradiance remains weak (r = 0.11), possibly due to localized dust events or temporal averaging that masks short-term attenuation effects. Furthermore, elevation demonstrates a mild inverse relationship with AOD (r = −0.13), consistent with reduced dust concentrations at higher altitudes due to gravitational settling and topographic shielding [57]. Lastly, elevation and water proximity exhibit a moderate positive correlation (r = 0.30), likely reflecting coastal topography where higher terrain is set back from the shoreline. Overall, the weak to moderate correlations suggest limited multi-collinearity among predictors, supporting their inclusion in multivariate modelling frameworks for hydrogen suitability without redundancy.

### 4.2 Discussion of Unsupervised Clustering Results

To address the absence of direct hydrogen yield data, this study employed an unsupervised learning approach using KMeans clustering to generate proxy suitability classes for green hydrogen production across nine cities. The clustering was based on a multidimensional feature space in the dataset. The algorithm partitioned the dataset into five optimal clusters, which were mapped to ordinal proxy classes labelled as "Very Low" to "Very High" suitability. Clusters ranked as "High" and "Very High" were characterized by consistently strong solar irradiance (mean > 6.0 kWh/m²/day), low AOD (mean < 0.3), and moderate elevation (< 400 m). In contrast, clusters associated with "Low" or "Very Low" suitability had high AOD (> 0.45), sparse vegetation cover (NDVI class ≤ 60), or proximity to mountainous or remote inland terrain, which may increase capital and operational expenditure for hydrogen infrastructure [58] [59]. The use of proxy classes allowed for the transformation of a continuous environmental landscape into an interpretable categorical framework for supervised learning and explainability modelling. The inclusion of the month feature further captured seasonal effects characteristic of gulf region dust and irradiance cycles [53]. This clustering method provides a scalable and explainable foundation for generating synthetic labels in data-scarce environments.

### 4.3 History Plots and Evaluation Metrics

Figure 5 shows the model history plot, which shows the model accuracy and loss over training epochs for both the training and test datasets to identify possible over-fitting or convergence issues. The plot shows an initial rapid increase in accuracy for both datasets, showing that the model quickly learned the data at about 50 epochs, after which the model reached optimal performance and was no longer making significant improvements. The chart also shows that the training and test accuracy curves are closely aligned, indicating minimal over-fitting, and therefore, the model generalizes well to unseen data. The smooth learning curve also suggests that the model hyper-parameters are well optimised

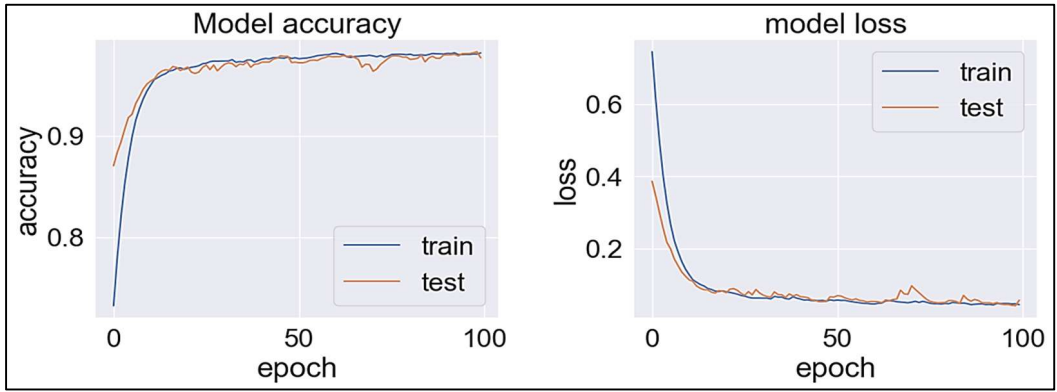

**Figure 5** History plot of model training and evaluation accuracy and loss

In addition to *accuracy*, the *precision*, *recall*, and *F1-score* metrics were used to evaluate the performance of the XGBoost model, as shown in Table 2 below.

**Table 2** Classification Report for the XGBoost model Predicting Proxy Class. .

|  | *Precision* | *Recall* | *F1-score* |
|---:|:---:|:---:|:---:|
| *Very Low* | 0.98 | 0.93 | 0.95 |
| *Low* | 0.97 | 0.98 | 0.98 |
| *Moderate* | 0.99 | 0.98 | 0.98 |
| *High* | 0.98 | 0.99 | 0.99 |
| *Very High* | 0.97 | 0.97 | 0.97 |
| *Accuracy* |  |  | 0.98 |
| *Macro Average* | 0.99 | 0.97 | 0.98 |
| *Weighted Average* | 0.98 | 0.97 | 0.98 |

### 4.4 SHAP-Based Feature Importance and Interpretability

Figure 6 below shows the multi-class SHAP summary plot and the corresponding mean absolute SHAP values shown on Table 3.

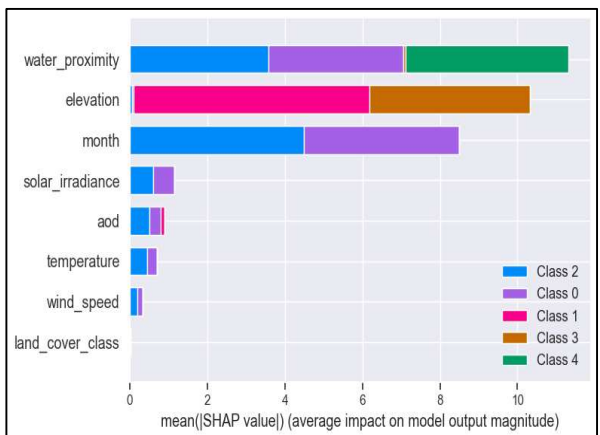

**Figure 6** SHAP Summary plot of Input Features Importance

**Table 2** Mean Absolute SHAP values of input Features

| *Feature* | *Mean |SHAP value|* |
|---:|:---:|
| *water_proximity* | 2.470891 |
| *elevation* | 2.376296 |
| *month* | 1.273216 |
| *solar_irradiance* | 0.396316 |
| *aod* | 0.280934 |
| *temperature* | 0.143496 |
| *wind_speed* | 0.113754 |
| *land_cover_class* | 0.005367 |

These results show that water_proximity, elevation, and month are the top three drivers influencing suitability classification. Specifically, water_proximity had the highest mean absolute SHAP value (2.26), which shows its significance in determining logistical and operational viability for hydrogen plant siting, particularly for desalination-integrated electrolysis projects in arid coastal environments [60] [61]). Likewise, elevation (2.07) emerged as a significant negative determinant, likely due to its association with infrastructure cost, pumping energy, and terrain inaccessibility [62]. The variable month (1.70) exhibited temporal influence across suitability classes, showing the seasonality of environmental conditions. In contrast, while solar_irradiance (0.23), AOD (0.18), and temperature (0.14) are physically critical to hydrogen production efficiency, their lower SHAP values suggest that their contribution to classification was less discriminative across the country. Also, wind_speed and land_cover_class had minimal SHAP impact, implying limited predictive utility in this context, or that their influence is indirectly captured by other correlated features such as month or elevation. These findings demonstrate the power of SHAP in model interpretability and align with prior studies that emphasize explainable AI as an important element of spatial decision-making in the energy sector [63] [64].

### 4.5 Composite Yield Index Computation

Building on the SHAP analysis, a composite site suitability index was constructed by using the mean absolute SHAP values as feature weights. Features were first scaled using min-max normalization to ensure comparability. Features like AOD, elevation, and water_proximity, which are negatively correlated with hydrogen yield and site suitability were inversely normalized (i.e., transformed to $1 - normalised\ values$) so that higher composite scores consistently reflect higher suitability. The resulting index provides a continuous ranking of potential yield and site suitability that is both interpretable and sensitive to feature importance as empirically derived from the model [64].

The distribution of the SHAP-guided Green Hydrogen Yield Composite Index and its derived categorical classes are presented in Figure 7. The continuous index was segmented into five ordinal classes as follow: "Very High" (SCI ≥ 5.4), "High" (4.4 ≤ SCI < 5.4), "Moderate" (3.4 ≤ SCI < 4.4), "Low" (2.4 ≤ SCI < 3.4), and "Very Low" (SCI < 2.4). This transformation enhances discrete interpretability levels for decision-making.

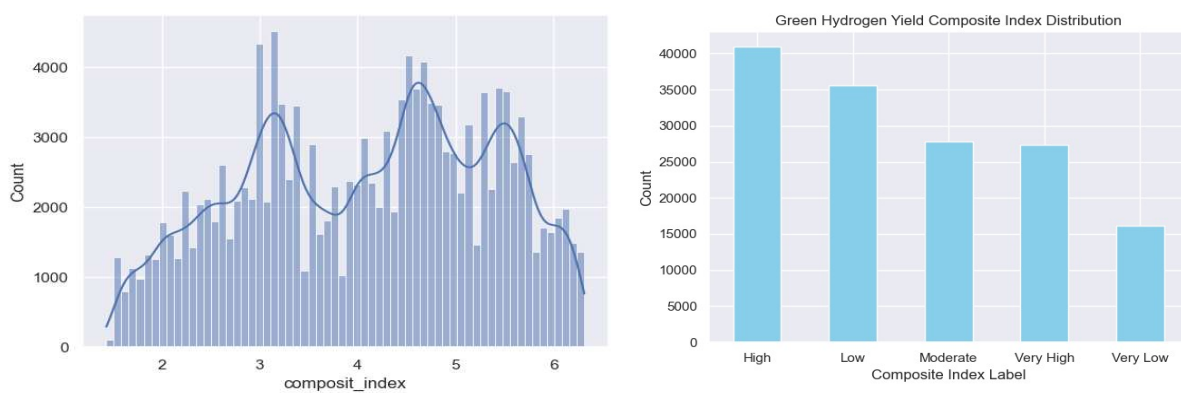

**Figure 7** Distribution of Composite Index for Green Hydrogen Yield and Site Suitability

The results reveal a skewed distribution with the majority of locations falling within the "High" (over 41,000 instances) and "Low" (over 36,000 instances) classes. This suggests a wide availability of technically viable sites across Oman, albeit with significant differentiation based on elevation and proximity to water bodies, variables that had the highest SHAP-derived weights. Surprisingly, the "Very High" class did not dominate the upper distribution tail, appearing in similar frequency to the "Moderate" group (27,000 each), indicating that only a fraction of the study area meets all optimal conditions simultaneously. The "Very Low" class comprised the smallest group (16,000), which may correspond to sites that are inland, elevated, or environmentally constrained (e.g., desert interiors or mountainous regions), and thus less suitable without substantial infrastructural investment. The class boundaries used in the logic function reflect a near-equal interval binning strategy over the index range, with an emphasis on enhancing ordinal clarity.

## 5. System Deployment

To translate the modelling insights into an operational decision-support tool, the entire SHAP-guided framework comprising the multivariable clustering engine, the ML classifier, the feature-explainability layer, and the composite Site Suitability Index was developed into a cloud-hosted, interactive dashboard (Figure 8). The application is built on a Python FastAPI back-end while the front-end is implemented in Streamlit and Plotly-Dash. The dashboard features a Scenario Builder which provides sliders and drop-downs allowing users to adjust variables and instantly visualise how these constraints impact the green hydrogen yield outcome. There is also an Explainability Widget which provides real-time insight on features ranking of each prediction and scenario simulation.

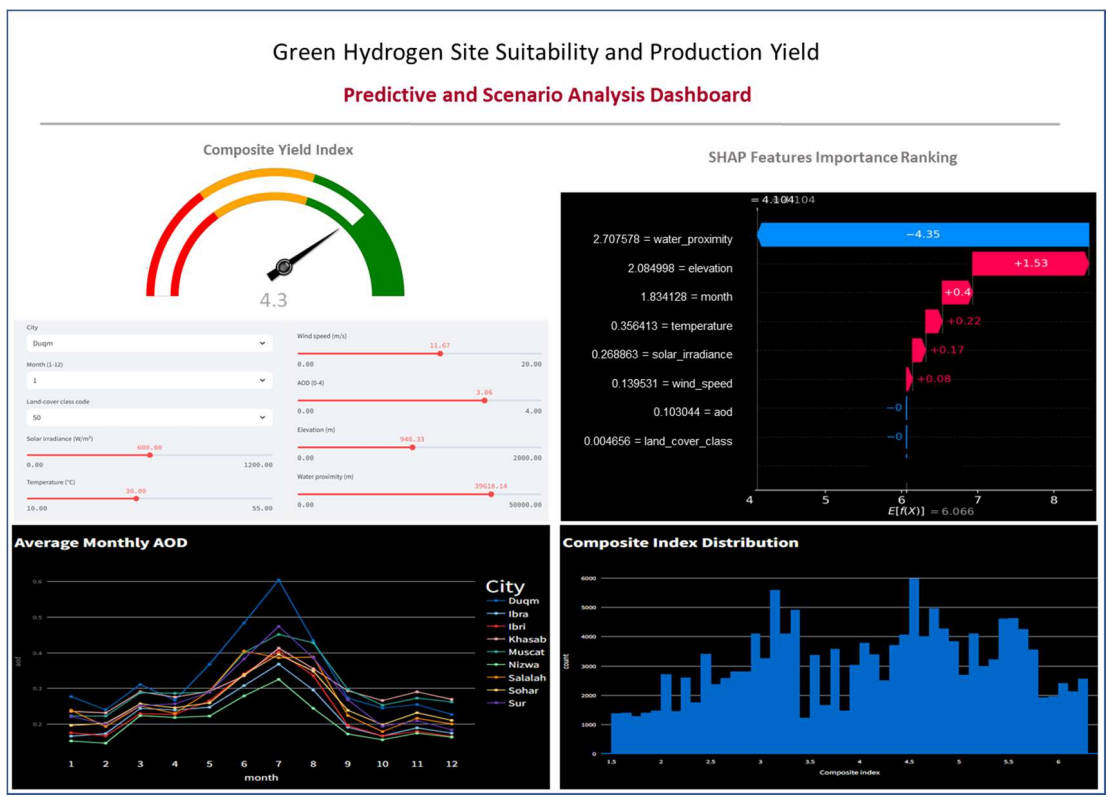

**Figure 8** Scenario Analysis and Predictive Analytics Dashboard for Green Hydrogen Yield and Site Suitability

# 6. Conclusion

This study presented a novel, explainable AI framework for evaluating green hydrogen site suitability in arid environments, with a specific focus on Oman. To enhance scalability and facilitate application in regions with limited or unavailable hydrogen yield data, an unsupervised clustering approach was adopted to derive proxy suitability classes based on key environmental, topographic, and proximity-related features. This was followed by the use of a supervised machine learning classifier to train the proxy classes, and SHAP algorithm was used to determine global feature importance and inform the construction of a composite Hydrogen Yield and Site Suitability Index. This index integrates variable weights derived from model-learned SHAP values, which provides a transparent data-driven alternative to conventional multi-criteria decision-making methods. The findings revealed that water proximity, elevation, and seasonality (month) are the most influential determinants of hydrogen yield and site suitability, surpassing even solar irradiance in predictive relevance due to their greater spatial variability. The research was developed into a dashboard which operationalizes the framework into an interactive tool for scenario analysis, geospatial exploration, and real-time decision-making. This work contributes a novel methodology by integrating supervised and unsupervised machine learning techniques, explainability tools, and geospatial analysis into a unified, operational framework aimed at supporting hydrogen transition strategies in regions constrained by limited data and climate-related challenges.

# 7. Industry Relevance and Policy Implications

The findings of this study offer significant implications for both industry stakeholders and policymakers seeking to accelerate the deployment of green hydrogen infrastructure in Oman and similar arid environments. From an industry perspective, the SHAP-guided composite index provides a scientifically grounded and data-driven basis for prioritizing investment zones, optimising site selection, and reducing preliminary exploration costs. Project developers and contractors can leverage the site suitability classifications to identify high-potential locations. Utility planners and infrastructure providers can gain valuable insights for aligning grid extension strategies, water sourcing, and pumping energy requirements with topographically favourable regions. Furthermore, the ability to show seasonality effects (e.g. mid-year spike in AOD) enhances its relevance for operations and maintenance scheduling, component specification, and performance forecasting for electrolyser and PV in dust-prone climates. In terms of the policy-level implications, by generating interpretable and spatially granular index, the framework enables evidence-based zoning and supports the formulation of hydrogen development corridors based on environmental and logistical realities. This can be important for processes such as permit streamlining, land use conflicts, and regulatory standards for site allocation. The high importance attributed to water proximity in the SHAP analysis shows the strategic necessity of aligning hydrogen deployment with national desalination planning, especially under Oman's increasing water stress and Vision 2040 sustainability targets [65]. Moreover, the integration of machine learning explainability enhances institutional transparency which facilitates trust and accountability in digital energy governance. This study demonstrates the nexus of data science and infrastructure delivery and also provides a transferable template for climate-resilient projects in the Gulf and other arid regions.

# 8. Suggestions for Future Work

Future research could extend this framework by incorporating economic and infrastructure variables such as grid proximity and road access, to support more comprehensive techno-economic assessments. Additionally, integrating climate projections and dynamic environmental data would enable the analysis of long-term suitability under climate change. Finally, future work may explore alternative explainability methods to validate assumptions and improve practical adoption.